# Primal-Dual Wasserstein GAN

**Mevlana Gemici**[†]  
mevlana.gemici@gmail.com

**Zeynep Akata**  
zeynepakata@gmail.com

**Max Welling**  
welling.max@gmail.com

## Abstract

We introduce *Primal-Dual* Wasserstein GAN—a new learning algorithm for building latent variable models of the data distribution based on the primal and the dual formulations of the optimal transport (OT) problem. We utilize the primal formulation to learn a flexible inference mechanism and to create an *optimal approximate coupling* between the data distribution and the generative model. In order to learn the generative model, we use the dual formulation and train the decoder adversarially through a critic network that is *regularized* by the approximate coupling obtained from the primal. Unlike previous methods that violate various properties of the *optimal critic*, we regularize the norm *and the direction* of the gradients of the critic function. Our model shares many of the desirable properties of auto-encoding models in terms of mode coverage and latent structure, while avoiding their undesirable averaging properties, e.g. their inability to capture sharp visual features when modeling real images. We compare our algorithm with several other generative modeling techniques that utilize Wasserstein distances on Fréchet Inception Distance (FID) and Inception Scores (IS).

## 1 Introduction

A prominent approach to unsupervised learning with deep latent variable models is the paradigm of Variational Autoencoders (VAEs, [1, 2]), which apply the principles of maximum likelihood estimation and variational posterior inference for learning. This paradigm provides a stable algorithmic framework for capturing high dimensional and complex data distributions with good generalization performance. However, it also requires generative models to have explicit densities and noise terms in the sample space to make inference possible. In VAEs, as well as in other auto-encoding based models, a combination of model mismatch and poor estimation of the posterior due to approximation/amortization gaps results in systematic biases in the learned distribution, e.g. undesirable averaging effects that are reflected in the samples produced.

The framework of Generative Adversarial Networks (GANS, [3]) is an antithesis to the paradigm of maximum likelihood estimation in several ways. Training of GANs requires neither an inference mechanism, nor a generative distribution that admits an explicit density to learn the parameters of the model. In GANs, estimation of the discrepancy between the data distribution and the generative model is accomplished through a *divergence approximator* [4] that is learned adversarially using independent samples from both distributions. This estimation is sound only at the non-parametric limit and the gap between the actual discrepancy and the estimated *lower* bound needs to be reduced through careful guidance of the functions selected to represent the divergence.

This work builds up on the theoretical analysis presented in [5, 6]. Our objective is to find a synthesis of the desirable qualities of the auto-encoding models and GANs using the theory of *Optimal Transport* [7] and statistical divergences associated with this framework, i.e. Wasserstein distances. Although auto-encoding models like VAEs have been combined with GANs in many different ways in recent literature (e.g. [8, 9, 10, 11, 12, 13]) in heuristic fashion, none of these succeed in bringing out the best of both paradigms and provide a strong theoretic justification. In this work, we bring together the two facets of the optimal transport problem by using the approximate auto-encoding solution to the primal problem for guiding the functions that represent the dual Wasserstein distance used for training the generative model.

---

[†]Corresponding author.

## 2 Notations and preliminaries

We will denote scalars and functions with lowercase italic letters (i.e. $f$), random variables and mappings with capital letters (i.e. $X, G$), and vectors with bold lowercase letters (i.e. $\mathbf{x}$). We use calligraphic letters (i.e. $\mathcal{F}, \mathcal{X}$) to denote sets. $\text{Prob}(\mathcal{X})$ denotes the class of all probability measures defined on $\mathcal{X}$. Each probability measure is indicated with letters $\pi$, $\tau$, $\mathbb{P}$ and $\mathbb{Q}$ with an appropriate subscript where necessary (i.e. $\pi_{Y|X}, \mathbb{P}_{\text{data}}, \mathbb{Q}_{Z|X}$). Greek letter $\lambda$ is reserved for (non-negative) trade-off parameters. $\mathfrak{supp}\,\mathbb{P}$ denotes the support of a probability measure.

Given a mapping $F \in \mathcal{F} : \mathcal{X} \to \mathcal{Y}$, the pushforward measure $F_\#\mu \in \text{Prob}(\mathcal{Y})$ is the distribution obtained by the deterministic transformation of the probability mass of $\mathbf{x} \sim \mu \in \text{Prob}(\mathcal{X})$ according to $\mathbf{y} = F(\mathbf{x})$. Given two probability measures, $\mathbb{P}_X \in \text{Prob}(\mathcal{X})$ and $\mathbb{P}_Y \in \text{Prob}(\mathcal{Y})$, $\pi \in \Pi(\mathbb{P}_X, \mathbb{P}_Y)$ indicates the set of all *couplings* or joint probability measures in the product space $\mathcal{X} \times \mathcal{Y}$ with specified marginals $\pi(\cdot, \mathcal{Y}) = \mathbb{P}_X$ and $\pi(\mathcal{X}, \cdot) = \mathbb{P}_Y$. We will refer to the product measure $\mathbb{P}_X \otimes \mathbb{P}_Y$ with independence $X \perp\!\!\!\perp Y$ as the *trivial coupling* and denote it as $\pi^\otimes \in \Pi(\mathbb{P}_X, \mathbb{P}_Y)$.

A *divergence* is defined as a function $d : \mathcal{A} \times \mathcal{A} \to \mathbb{R}^+$, such that $d(\mathbf{a}, \mathbf{b}) = 0$ iff $\mathbf{a} = \mathbf{b}$ (non-negativity and identity of indiscernibles). A *metric* (or distance) is a divergence with symmetry $d(\mathbf{a}, \mathbf{b}) = d(\mathbf{b}, \mathbf{a})$ and subadditivity $d(\mathbf{a}, \mathbf{c}) \leq d(\mathbf{a}, \mathbf{b}) + d(\mathbf{b}, \mathbf{c})$ for all $\mathbf{a}, \mathbf{b}, \mathbf{c} \in \mathcal{A}$. A *statistical* (or probability) divergence $\mathfrak{D}(\mu\|\nu) : \text{Prob}(\mathcal{X}) \times \text{Prob}(\mathcal{X}) \to \mathbb{R}^+$ measures the discrepancy between two probability measures defined over the same set. Wasserstein distances, which is the central topic of this work, is a particular example of an integral probability metric (IPM), which is a class of statistical divergences. In Appendix D, we summarize the benefits of using IPMs in generative modeling in comparison to more commonly known $f$-divergences, such as KL divergence used in maximum likelihood estimation.

## 3 Primal

In this work, we are interested in Wasserstein distances, a family of probability metrics based on the theory of Optimal Transport. The optimal transport problem provides a way to measure the distance between two distributions from the perspective of transportation of probability mass. The *primal* of the optimal transport problem formulated by Kantorovich [7] is given as,

$$\text{OT}_c(\mathbb{P}_{\text{data}}, \mathbb{P}_G) = \inf_{\pi \in \Pi(\mathbb{P}_{\text{data}}, \mathbb{P}_G)} \int_{\mathcal{X} \times \mathcal{Y}} c(\mathbf{x}, \mathbf{y})\, d\pi(\mathbf{x}, \mathbf{y}) = \inf_{\pi \in \Pi(\mathbb{P}_{\text{data}}, \mathbb{P}_G)} \mathbb{E}_{(\mathbf{x}, \mathbf{y}) \sim \pi}\left[c(\mathbf{x}, \mathbf{y})\right] \quad (1)$$

where $c : \mathcal{X} \times \mathcal{Y} \to \mathbb{R} \cup \{+\infty\}$ is a given cost function for transporting a unit of mass and the infimum is over all *couplings* between $\mathbb{P}_{\text{data}}$ and $\mathbb{P}_G$. The objective in optimal transport is to find an *optimal coupling* $\pi^* \in \Pi(\mathbb{P}_{\text{data}}, \mathbb{P}_G)$ such that the aggregate transport cost of moving (probability) mass from $\mathbb{P}_{\text{data}}$ to $\mathbb{P}_G$ (and vice versa if $c$ is symmetrical) is minimized. Given $\mathbb{P}_{\text{data}}$ and $\mathbb{P}_G$ are defined on the same (Polish) metric space $(\mathcal{X}, d)$, the $p$-Wasserstein distance between them is

$$\mathfrak{D}_p^W(\mathbb{P}_{\text{data}} \| \mathbb{P}_G) = \text{OT}_{d^p}(\mathbb{P}_{\text{data}}, \mathbb{P}_G)^{1/p} = \left(\inf_{\pi \in \Pi(\mathbb{P}_{\text{data}}, \mathbb{P}_G)} \mathbb{E}_{(\mathbf{x}, \mathbf{y}) \sim \pi}\left[d(\mathbf{x}, \mathbf{y})^p\right]\right)^{1/p}. \quad (2)$$

Note that $\mathfrak{D}_p^W(\mathbb{P}_{\text{data}} \| \mathbb{P}_G)$ is a metric proper in $\text{Prob}(\mathcal{X})$ for $p \in [1, \infty)$. In this work, we focus on 1-Wasserstein distances for Euclidean metric spaces, i.e. $d(\mathbf{x}, \mathbf{y}) = \|\mathbf{x} - \mathbf{y}\|$.

### 3.1 Reparameterization

Searching over the space of couplings, $\Pi(\mathbb{P}_{\text{data}}, \mathbb{P}_G)$, is a difficult task when $\mathbb{P}_{\text{data}}$ and $\mathbb{P}_G$ are complex distributions in high dimensional spaces. In general, maintaining the marginal constraints, $\pi(\cdot, \mathcal{X}) = \mathbb{P}_X$ and $\pi(\mathcal{X}, \cdot) = \mathbb{P}_G$, of a given joint distribution $\pi \in \text{Prob}(\mathcal{X} \times \mathcal{X})$ during optimization is challenging.

Consider the special case where the generative distribution $\mathbb{P}_G$ can be specified as a latent variable model that maps latent codes from a simple prior distribution $\mathbf{z} \sim \mathbb{P}_Z \in \text{Prob}(\mathcal{Z})$ to the sample space $\mathcal{X}$ through a deterministic decoder $G$, i.e. $\mathbf{y} = G(\mathbf{z})$ and $\mathbb{P}_G = G_\#\mathbb{P}_Z$. In this case, it is possible to *reparameterize* the constrained optimization problem in Eq. 1 as ([6], see Fig. 1),

$$\text{OT}_{\|\cdot\|}(\mathbb{P}_{\text{data}}, \mathbb{P}_G) = \inf_{\tau \in \Pi(\mathbb{P}_{\text{data}}, \mathbb{P}_Z)} \mathbb{E}_{(\mathbf{x}, \mathbf{z}) \sim \tau} \|\mathbf{x} - G(\mathbf{z})\| = \inf_{\substack{\mathbb{Q}_{Z|X} \text{ s.t.} \\ \mathbb{Q}_Z = \mathbb{P}_Z}} \mathbb{E}_{\substack{\mathbf{x} \sim \mathbb{P}_{\text{data}} \\ \mathbf{z} \sim \mathbb{Q}_{Z|X=\mathbf{x}}}} \|\mathbf{x} - G(\mathbf{z})\| \quad (3)$$

which is over all conditional distributions $\mathbb{Q}_{Z|X}$ such that the *aggregated posterior* $\mathbb{Q}_Z$, which is the marginal distribution over $\mathcal{Z}$ when $\mathbf{x} \sim \mathbb{P}_{\text{data}}$ and $\mathbf{z} \sim \mathbb{Q}_{Z|X=\mathbf{x}}$, is equivalent to the prior, i.e. $\mathbb{Q}_Z = \mathbb{P}_Z$.

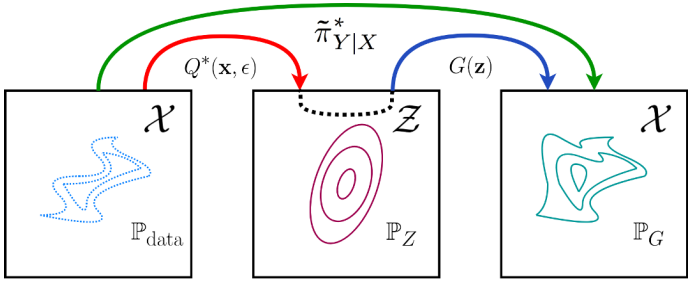

**Figure 1: Reparameterization of couplings**: Given a latent variable model $\mathbb{P}_G$ specified by prior $\mathbb{P}_Z$ and a deterministic map $G(\mathbf{z})$ (shown in blue), *optimal POT coupling* $\tilde{\pi}^*$ (shown in green) can be *reparameterized* by combining $G(\mathbf{z})$ with the optimal encoder $Q^*(\mathbf{x}, \epsilon)$ according to Eq. 5. Sampler $Q^*$ parameterizes a probabilistic encoder $\mathbb{Q}^*_{Z|X}$ (shown in red) that maps $\mathbb{P}_{\text{data}}$ to $\mathbb{P}_Z$.

In effect, the reparameterization removes from the marginal constraints, the decoder of the generative model $G$, a highly complex and nonlinear mapping between the latent space and the sample space, and transfers it inside the expectation. Instead of matching $\mathbb{P}_\mathbb{G}$, this simplified constrained optimization problem requires us to match only the prior $\mathbb{P}_Z$, which is defined over a lower dimensional space and corresponds to a much simpler distribution in general.

### 3.2 Penalized Optimal Transport

The constrained optimization problem in Eq. 3 can be relaxed by introducing a divergence between the prior $\mathbb{P}_Z$ and the aggregated posterior $\mathbb{Q}_Z$ as a penalty for violating the constraint $\mathbb{Q}_Z = \mathbb{P}_Z$. This is referred to as the *penalized optimal transport* objective [6, 14]:

$$\text{POT}_{\|\cdot\|}(\mathbb{P}_{\text{data}}, \mathbb{P}_G) = \inf_{\mathbb{Q}_{Z|X}} \mathbb{E}_{\substack{\mathbf{x} \sim \mathbb{P}_{\text{data}} \\ \mathbf{z} \sim \mathbb{Q}_{Z|X=\mathbf{x}}}} \|\mathbf{x} - G(\mathbf{z})\| + \lambda_Z \cdot \mathfrak{D}(\mathbb{Q}_Z \| \mathbb{P}_Z) \quad (4)$$

The divergence $\mathfrak{D}$ can be chosen based on several considerations. In the case of an empirical data distribution $\mathbb{P}_{\text{data}}$ that can be viewed as a uniform mixture of Dirac measures, the density of the aggregated posterior (assuming it exists) cannot be efficiently computed for large datasets. Therefore, a direct application of divergences that require computation of this density, such as $f$-divergences, is not tractable. Instead, a divergence computation that relies solely on samples from the distributions would be much more suitable for this task (e.g. estimating integral probability metrics such as MMDs or GAN based divergence approximators using independent samples).

Since its density does not need to be tractable in this case, the *probabilistic encoder* $\mathbb{Q}_{Z|X}$ can be parameterized by highly flexible function approximators without a fixed probability class (e.g. unlike Gaussian encoders in regular VAEs [1]). This can be accomplished by introducing a random noise component from a fixed distribution $\epsilon \sim \mathbb{N}(0, I)$ (i.e. a multivariate Normal distribution), and parameterizing the encoder as a flexible deterministic sampler $\mathbf{z} = Q(\mathbf{x}, \epsilon)$ where $Q \in \mathcal{Q} : \mathcal{X} \cup \mathcal{E} \to \mathcal{Z}$:

$$\text{POT}_{\|\cdot\|}(\mathbb{P}_{\text{data}}, \mathbb{P}_G) = \min_{Q \in \mathcal{Q}} \mathbb{E}_{\substack{\mathbf{x} \sim \mathbb{P}_{\text{data}} \\ \epsilon \sim \mathbb{N}(0,I)}} \|\mathbf{x} - G(Q(\mathbf{x}, \epsilon))\| + \lambda_Z \cdot \mathfrak{D}(Q_\# \mathbb{P}_{\text{data}} \otimes \mathbb{N}(0, I) \| \mathbb{P}_Z) \quad (5)$$

The sampler $Q$ can represent any conditional distribution $\mathbb{Q}_{Z|X}$ at the non-parametric limit, i.e. when $\mathcal{Q}$ is a universal function approximator class, and when the intrinsic dimensionality of the random noise component is (at least) as large as the dimensionality of $\mathcal{Z}$.

### 3.3 Optimal POT Couplings

Minimizing the *penalized optimal transport* objective in Eq. 5 corresponds to learning a flexible inference mechanism for a *fixed* generative model $\mathbb{P}_G$ through a probabilistic encoder $\mathbb{Q}_{Z|X}$ that is parameterized by the sampler $Q$ and regularized by the prior over latent codes $\mathbb{P}_Z$. For a given $\lambda_Z$, the optimal sampler $Q^*$ of Eq. 5, along with the decoder $G$, parameterizes an *optimal POT coupling* $\tilde{\pi}^*$ between $\mathbb{P}_{\text{data}}$ and $\mathbb{P}_G$ (see Figure 1). Under certain regularity conditions on the divergence $\mathfrak{D}$, the optimal encoder $\mathbb{Q}^*_{Z|X}$ parameterized by $Q^*$ is confined into the feasible region[2] (i.e. $\mathbb{Q}^*_Z = \mathbb{P}_Z$) as $\lambda_Z \to \infty$; and results in a valid coupling, i.e. $\tilde{\pi}^* \in \Pi(\mathbb{P}_{\text{data}}, \mathbb{P}_G)$.

Samples from this (possibly approximate) coupling, $(\mathbf{x}, \mathbf{y}) \sim \tilde{\pi}^*$, can be obtained by first sampling a point from the data distribution $\mathbf{x} \sim \mathbb{P}_{\text{data}}$ and then reconstructing it under $Q^*$ and $G$, i.e. $\mathbf{y} = G(Q^*(\mathbf{x}, \epsilon))$, which gives its coupled pair (see Figure 1). In the dual problem, we use samples from $\tilde{\pi}^*$ as a source of regularization for *learning* the generative model.

---

[2]Equivalent to optimization with constraint $\mathfrak{D}(\mathbb{Q}_Z \| \mathbb{P}_Z) \le r_{\lambda_Z}$, where $r_{\lambda_Z} \to 0$ as $\lambda_Z \to \infty$.

## 4 Dual

The duality theorem for the optimal transport problem provide us with a different perspective. While in the primal problem, the central notion is the *minimal cost*; in the dual problem, it is *competitive price*. In the following, we present a special case of the Kantorovich duality [7] that is relevant to our setting:

**Theorem 1 (Kantorovich-Rubenstein Duality).** *Let $(\mathcal{X}, d)$ be a compact metric space and $\mu$ and $\nu$ be two (probability) measures in $\mathrm{Prob}(\mathcal{X})$. If the unit transport cost is the distance in this metric space, $c(\mathbf{x}, \mathbf{y}) = d(\mathbf{x}, \mathbf{y})$, then,*

*(i) there is duality,*

$$\inf_{\pi \in \Pi(\mu,\nu)} \int_{\mathcal{X} \times \mathcal{X}} d(\mathbf{x}, \mathbf{y}) \, d\pi(\mathbf{x}, \mathbf{y}) = \sup_{f \in \mathcal{F}_{\mathrm{Lip}}} \int_{\mathcal{X}} f(\mathbf{x}) \, d\mu(\mathbf{x}) - \int_{\mathcal{X}} f(\mathbf{y}) \, d\nu(\mathbf{y}) \quad (6)$$

*where the supremum is over the class of all bounded 1-Lipschitz[3] functions $f : \mathcal{X} \to \mathbb{R}$.*

*(ii) the following two statements are equivalent:*

*(a) $\pi^*$ is an optimal coupling for the primal;*
*(b) For any optimal potential function for the dual, $f^* \in \mathcal{F}_{\mathrm{Lip}}$, $f^*(\mathbf{x}) - f^*(\mathbf{y}) = d(\mathbf{x}, \mathbf{y})$ is satisfied $\pi^*$-almost surely, i.e. $\mathbb{E}_{(\mathbf{x},\mathbf{y}) \sim \pi^*}\left[\mathbb{1}[f^*(\mathbf{x}) - f^*(\mathbf{y}) = d(\mathbf{x}, \mathbf{y})]\right] = 1.$*

Applying Theorem 1 to our setting, the dual optimal transport problem and 1-Wasserstein distance between $\mathbb{P}_{\mathrm{data}}$ and $\mathbb{P}_G$ can now be written as [5]:

$$\mathrm{OT}_d(\mathbb{P}_{\mathrm{data}}, \mathbb{P}_G) = \sup_{f \in \mathcal{F}_{\mathrm{Lip}}} \mathbb{E}_{\mathbf{x} \sim \mathbb{P}_{\mathrm{data}}}[f(\mathbf{x})] - \mathbb{E}_{\mathbf{y} \sim \mathbb{P}_G}[f(\mathbf{y})] = \sup_{f \in \mathcal{F}_{\mathrm{Lip}}} \mathbb{E}_{\mathbf{x} \sim \mathbb{P}_{\mathrm{data}}}[f(\mathbf{x})] - \mathbb{E}_{\mathbf{z} \sim \mathbb{P}_Z}[f(G(\mathbf{z}))] \quad (7)$$

The objective in the dual problem is to find an optimal *potential* $f^* \in \mathcal{F}_{\mathrm{Lip}}$ such that it assigns the maximum aggregate value (price) to points that are concentrated on $\mathbb{P}_{\mathrm{data}}$ and the minimum aggregate value to points that are concentrated on $\mathbb{P}_G$. The Lipschitz constraint, however, grounds the potential function such that absolute differences between individual values that it assigns over all $\mathcal{X}$ are bounded by a constraint of similarity, where similarity is measured by distance in this metric space. Therefore, potential functions depend on the metric space in which the data points lie in and are forced to assign similar values to points that are close to each other in this metric space.

In Wasserstein GANs (WGAN, [5]), the potential $f \in \mathcal{F}_{\mathrm{Lip}}$ is represented by a *critic* network with parameters that lie within a compact space $[-b, b]$ to ensure a $k$-Lipschitz constraint is satisfied. Once the parameters of the critic network $f$ is learned sufficiently well by maximizing Eq. 7, the dual distance based on this critic network is minimized to learn the parameters of the decoder $G$. However, parameterizing the critic network such that it cannot lie outside the feasible set during any part of the optimization process can over-constrain it and make WGANs prohibitively ineffective at reaching an optimal potential (referred to as the *capacity underuse* problem in [15]).

### 4.1 Couplings and Potentials at Optimality

The optimal transport problem and its dual formulation have extensively studied geometric properties that connect the geodesics (or shortest paths) of the metric space $(\mathcal{X}, d)$, optimal couplings $\pi^*$ and optimal potential functions $f^*$ together, e.g. the general statement given in Theorem 1 (ii) that is valid for any metric space. In this section, we examine a corollary to Theorem 1 (ii) that applies to *Euclidean* metric spaces [15] (see Appendix A for a proof):

**Corollary 1.** *Let $(\mathcal{X}, d(\mathbf{x}, \mathbf{y}) = \|\mathbf{x} - \mathbf{y}\|)$ be a compact metric space and the unit transport cost is the distance in this metric space, $c(\mathbf{x}, \mathbf{y}) = \|\mathbf{x} - \mathbf{y}\|$. Given $\pi^*$ is an optimal coupling for the primal problem and $f^*$ is an optimal potential for the dual problem, if $(\mathbf{x}, \mathbf{y}) \in \mathrm{supp}\, \pi^*$ with $\mathbf{x}$ and $\mathbf{y}$ distinct, $\mathbf{x}_t = (1-t) \cdot \mathbf{x} + t \cdot \mathbf{y}$ with $0 \leq t \leq 1$, and $f^*$ is differentiable at $\mathbf{x}_t$[4], then, $\nabla f^*(\mathbf{x}_t) = \mathbf{x} - \mathbf{y} / \|\mathbf{x} - \mathbf{y}\|$ is satisfied $\pi^*$-almost surely, i.e. $\mathbb{E}_{\substack{(\mathbf{x},\mathbf{y}) \sim \pi^* \\ t \sim \mathrm{Unif}[0,1]}}\left[\mathbb{1}[\nabla f^*(\mathbf{x}_t) = \mathbf{x} - \mathbf{y} / \|\mathbf{x} - \mathbf{y}\|]\right] = 1.$*

---

[3](The best) Lipschitz constant of a function is defined with respect to a specific metric space $(\mathcal{X}, d)$ as $\|f\|_L = \sup_{\mathbf{x}, \mathbf{y} \in \mathcal{X}} \{|f(\mathbf{y}) - f(\mathbf{x})| / d(\mathbf{x}, \mathbf{y})\}$. 1-Lipschitz functions, $f \in \mathcal{F}_{\mathrm{Lip}}$, satisfy $\|f\|_L \leq 1$.

[4]We write $\nabla f(\mathbf{x})$ to indicate the gradient of $f$ w.r.t its input instead of $\nabla_{\mathbf{x}} f(\mathbf{x})$ to simplify the notation.

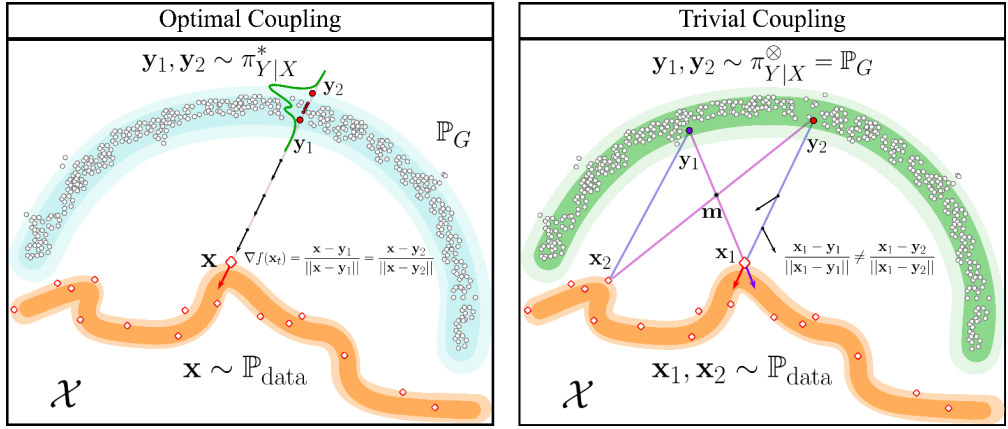

**Figure 2: Optimal couplings $\pi^*$ and optimal potentials $f^*$: Left**: We illustrate the conditional distribution $Y|X$ for an optimal coupling, $\pi^*_{Y|X}$ (in green). For an optimal potential $f^*$, gradient norms on the line segments between pairs $(\mathbf{x}, \mathbf{y}_1)$ and $(\mathbf{x}, \mathbf{y}_2)$ are always *one* and the gradient direction (black arrows) is *consistently* the unit vector pointing towards $\mathbf{x}$ from $\mathbf{y}_1$ and $\mathbf{y}_2$ (Corollary 1). Assuming it exists, this includes the gradient at $\mathbf{x}$ (red arrow), i.e. $\nabla f^*(\mathbf{x})$. In this case, $\mathbf{y}_1$ and $\mathbf{y}_2$ (and the entire $\mathrm{supp}\,\pi^*_{Y|X}$) must lie on the same half-line (Proposition 2). **Right**: We illustrate the conditional distribution $Y|X$ for the trivial coupling, $\pi^{\otimes}_{Y|X} = \mathbb{P}_G$ (in green). Corollary 1 does not apply to (all) line segments associated with $\pi^{\otimes}$, which significantly weakens the theoretical justification given for the regularization scheme used in WGAN-GPs.

Note that this property determines the norm *and* the direction of the gradients of $f^*$ on line segments between the coupled pairs $(\mathbf{x}, \mathbf{y})$ for any optimal coupling $\pi^*$. Authors of [15] use this property of optimal potentials to motivate their choice of regularization for the critic network in their model, *WGAN with Gradient Penalty* (WGAN-GP, [15]):

$$\max_{f \in \mathcal{F}} \mathbb{E}_{\mathbf{x} \sim \mathbb{P}_{\text{data}}}[f(\mathbf{x})] - \mathbb{E}_{\mathbf{z} \sim \mathbb{P}_Z}[f(G(\mathbf{z}))] + \lambda_f \cdot \mathbb{E}_{\substack{(\mathbf{x},\mathbf{y}) \sim \pi^{\otimes} \\ t \sim \text{Unif}[0,1]}}\left[\left(\|\nabla f(\mathbf{x}_t)\| - 1\right)^2\right] \quad (8)$$

WGAN-GP objective encourages the *norm* of the gradients of the critic $f$ to be close to $1$ on certain line segments in $\mathcal{X}$. Unlike Corollary 1, however, WGAN-GP uses samples from the *trivial coupling* $(\mathbf{x}, \mathbf{y}) \sim \pi^{\otimes}$ to determine the line segments that the penalty in Eq. 8 is evaluated at, where the endpoints $\mathbf{x}$ and $\mathbf{y}$ are independently chosen from $\mathbb{P}_{\text{data}}$ and $\mathbb{P}_G$ respectively. This is a practical choice that stems from not having access to sample pairs from an optimal coupling $(\mathbf{x}, \mathbf{y}) \sim \pi^*$. Note that gradient *directions* are left unregularized since $\mathbf{x}-\mathbf{y}/\|\mathbf{x}-\mathbf{y}\|$ would be a random direction for samples from the trivial coupling, which is the highest entropy (most random) coupling in $\Pi(\mathbb{P}_{\text{data}}, \mathbb{P}_G)$ [16].

In general, Corollary 1 does not apply to samples from the trivial coupling since it is guaranteed that $\mathrm{supp}\,\pi^{\otimes} \neq \mathrm{supp}\,\pi^*$ under mild conditions. We introduce the following set of propositions to outline some of the common cases where the trivial coupling cannot be used to obtain samples from an optimal coupling (see Fig. 2 for an intuition and Appendix A for proofs):

**Monge-Mather shortening principle, [7].** *No optimal coupling $\pi^*$ can have on its support two distinct pairs $(\mathbf{x}_1, \mathbf{y}_1)$ and $(\mathbf{x}_2, \mathbf{y}_2)$ such that the line segments associated with them intersect at an interior point, $\mathbf{m} = (1-t_1) \cdot \mathbf{x}_1 + t_1 \cdot \mathbf{y}_1 = (1-t_2) \cdot \mathbf{x}_2 + t_2 \cdot \mathbf{y}_2$ where $t_1, t_2 \in (0,1)$ (see Fig. 2).*

**Proposition 1.** *Given a neighborhood exists around any point $\mathbf{y} \in \mathrm{supp}\,\mathbb{P}_G$ along $\pm(\mathbf{x}_2-\mathbf{x}_1)/\|\mathbf{x}_2-\mathbf{x}_1\|$ for any two points $\mathbf{x}_1, \mathbf{x}_2 \in \mathrm{supp}\,\mathbb{P}_{\text{data}}$, then $\mathrm{supp}\,\pi^{\otimes} \neq \mathrm{supp}\,\pi^*$.*

**Proposition 2.** *Given $(\mathbf{x}, \mathbf{y}) \in \mathrm{supp}\,\pi^*$, and $f^*$ is differentiable at $\mathbf{x}$, then $\mathrm{supp}\,\pi^*_{Y|X=\mathbf{x}}$ is confined to (a subset of) the half line given by $\mathbf{x} + t \cdot (\mathbf{y}-\mathbf{x}/\|\mathbf{y}-\mathbf{x}\|)$ where $t \in [0,\infty)$.*

**Corollary 2.** *Given there exists a point $\mathbf{x} \in \mathrm{supp}\,\mathbb{P}_{\text{data}}$ where $f^*$ is differentiable at $\mathbf{x}$ and $\mathrm{supp}\,\mathbb{P}_G$ does not lie on a single line, then $\mathrm{supp}\,\pi^{\otimes} \neq \mathrm{supp}\,\pi^*$.*

These results weaken the theoretical justification given for the regularization scheme used in WGAN-GPs, which in practice has been utilized even in algorithms that are not directly connected to Wasserstein distances (e.g. [17]). Although regularization of gradient norms on line segments associated with the trivial coupling may have practical benefits in general, it is difficult to firmly place it in the optimal transport theory (e.g. [18] suggests a connection to Sobolev IPM as an alternative characterization).

In this work, we propose a gradient *vector* penalty that guides the training of the critic network such that it satisfies Corollary 1 on line segments associated with optimal POT couplings $\tilde{\pi}^*$ defined in Section 3.3, i.e. a much better approximation to optimal couplings than the trivial coupling:

$$\max_{f \in \mathcal{F}} \mathbb{E}_{\mathbf{x} \sim \mathbb{P}_{\text{data}}}[f(\mathbf{x})] - \mathbb{E}_{\mathbf{z} \sim \mathbb{P}_Z}[f(G(\mathbf{z}))] - \lambda_f \cdot \mathbb{E}_{\substack{(\mathbf{x},\mathbf{y}) \sim \tilde{\pi}^* \\ t \sim \text{Unif}[0,1]}} \left\| \nabla f(\mathbf{x}_t) - \frac{\mathbf{x} - \mathbf{y}}{\|\mathbf{x} - \mathbf{y}\|} \right\|^2 \quad (9)$$

Note that unlike the gradient *norm* penalty used in WGAN-GPs, we regularize gradient norms *and directions* since directions provided by the pairs of samples from $\tilde{\pi}^*$ are not random and closely approximate pairings from optimal couplings.

### 4.2 Auto-encoding Critic Networks

As we outlined in Section 3, optimal couplings interlink samples from $\mathbb{P}_{\text{data}}$ with samples from $\mathbb{P}_G$ in an auto-encoding manner when latent codes constitute a part of the generative model. However, the dual problem as formulated in Eq. 7 makes no direct reference to couplings aside from the solution at the optimality. Only at the optimality, properties of potentials are connected to optimal couplings as we have shown in Section 4.1.

In the following, we rearrange the dual problem such that samples from a (possibly sub-optimal) coupling, $\pi \in \Pi(\mathbb{P}_{\text{data}}, \mathbb{P}_G)$, are used in computing the dual Wasserstein distance along with independent samples from $\mathbb{P}_Z$ (see Appendix B for a derivation):

$$\text{OT}_d(\mathbb{P}_{\text{data}}, \mathbb{P}_G) = \sup_{f \in \mathcal{F}_{\text{Lip}}} \mathbb{E}_{(\mathbf{x},\mathbf{y}) \sim \pi}[f(\mathbf{x}) - \lambda_{\text{mix}} \cdot f(\mathbf{y})] - (1 - \lambda_{\text{mix}}) \cdot \mathbb{E}_{\mathbf{z} \sim \mathbb{P}_Z}[f(G(\mathbf{z}))]$$

$$= \sup_{f \in \mathcal{F}_{\text{Lip}}} \mathbb{E}_{\mathbf{x} \sim \mathbb{P}_{\text{data}}}\left[f(\mathbf{x}) - \lambda_{\text{mix}} \cdot \mathbb{E}_{\mathbf{z} \sim \mathbb{Q}_{Z|X=\mathbf{x}}}[f(G(\mathbf{z}))]\right] - (1 - \lambda_{\text{mix}}) \cdot \mathbb{E}_{\mathbf{z} \sim \mathbb{P}_Z}[f(G(\mathbf{z}))] \quad (10)$$

Eq. 10 is valid for any conditional distribution $\mathbb{Q}_{Z|X}$ whose *aggregated posterior* is equal to the prior, i.e. $\mathbb{Q}_Z = \mathbb{P}_Z$, as defined previously in Section 3.1. In the following proposition, we highlight the connection between the gradient of the first expectation term in Eq. 10 and the gradient of an auto-encoder with respect to parameters of the decoder when the potential function $f$ have certain properties (see Appendix C for a proof).

**Proposition 3.** *Assume $\theta$ is a set of variables that parameterizes the decoder $G$, $\mathbf{x} \sim \mathbb{P}_{\text{data}}$ is any given data sample and $f \in \mathcal{F}_{\text{Lip}}$ is a (possibly sub-optimal) potential function that $\mathbb{Q}_{Z|X=\mathbf{x}}$-almost surely satisfies $\nabla f(G(\mathbf{z})) = \mathbf{x} - G(\mathbf{z}) / \|\mathbf{x} - G(\mathbf{z})\|$ where $\mathbb{Q}_{Z|X}$ is any conditional distribution. Then, the following statement is true:*

$$\nabla_\theta \mathbb{E}_{\mathbf{x} \sim \mathbb{P}_{\text{data}}}\left[f(\mathbf{x}) - \lambda_{\text{mix}} \cdot \mathbb{E}_{\mathbf{z} \sim \mathbb{Q}_{Z|X=\mathbf{x}}}[f(G(\mathbf{z}))]\right] = \lambda_{\text{mix}} \cdot \nabla_\theta \mathbb{E}_{\mathbf{x} \sim \mathbb{P}_{\text{data}}}\left[\mathbb{E}_{\mathbf{z} \sim \mathbb{Q}_{Z|X=\mathbf{x}}}\|\mathbf{x} - G(\mathbf{z})\|\right].$$

When a critic network is trained with the gradient vector penalty in Eq. 9, it also approximately satisfies the condition in Proposition 3 for $\mathbb{Q}^*_{Z|X}$ associated with $\tilde{\pi}^*$. In this case, if $\tilde{\pi}^*$ is chosen as the coupling in Eq. 10, the first term of Eq. 10 will directly contribute to auto-encoding of the data distribution if the decoder of the generative model is trained to minimize it with a gradient based method.

## 5 Primal-Dual Wasserstein GAN

In this section, we introduce *Primal-Dual* Wasserstein GAN, a new algorithm for training an auto-encoding latent variable model of the data distribution that combines the primal and the dual formulations of the optimal transport problem. The main principle of our algorithm is to estimate the dual Wasserstein distance as formulated in Section 4.2 through a critic network that is regularized by the gradient properties of optimal potentials given in Section 4.1; and to minimize this distance by optimizing the parameters of the decoder to learn the generative model. The objective of the critic can be summarized as follows:

$$\max_f \mathbb{E}_{(\mathbf{x},\mathbf{y}) \sim \tilde{\pi}^*}[f(\mathbf{x}) - \lambda_{\text{mix}} \cdot f(\mathbf{y})] - (1 - \lambda_{\text{mix}}) \cdot \mathbb{E}_{\mathbf{z} \sim \mathbb{P}_Z}[f(G(\mathbf{z}))] - \lambda_f \cdot \mathbb{E}_{\substack{(\mathbf{x},\mathbf{y}) \sim \tilde{\pi}^* \\ t \sim \text{Unif}[0,1]}} \left\| \nabla f(\mathbf{x}_t) - \frac{\mathbf{x} - \mathbf{y}}{\|\mathbf{x} - \mathbf{y}\|} \right\|^2 \quad (11)$$

In PD-WGANs, we train an encoder network $Q$ that is used both for inference of latent variables and for obtaining samples from the optimal POT coupling; a critic network $f$ for computing the dual Wasserstein distance between the data distribution $\mathbb{P}_{\text{data}}$ and the generative model $\mathbb{P}_G$; and a decoder $G$ that parameterizes the generative model.

The objectives for each network in PD-WGANs are given as follows:

$$Q^* = \underset{Q}{\arg\min} \underset{\substack{\mathbf{x}\sim\mathbb{P}_{\text{data}} \\ \epsilon\sim\mathbb{N}(0,I)}}{\mathbb{E}} \|\mathbf{x} - G(Q(\mathbf{x},\epsilon))\| + \lambda_Z \cdot \mathfrak{D}(Q_\#\mathbb{P}_{\text{data}} \otimes \mathbb{N}(0,I) \| \mathbb{P}_Z) \quad (12)$$

$$f^* = \underset{f}{\arg\max} \underset{\substack{\mathbf{x}\sim\mathbb{P}_{\text{data}} \\ \epsilon\sim\mathbb{N}(0,I)}}{\mathbb{E}} \left[f(\mathbf{x}) - \lambda_{\text{mix}} \cdot f(G(Q^*(\mathbf{x},\epsilon)))\right] - (1-\lambda_{\text{mix}}) \cdot \underset{\mathbf{z}\sim\mathbb{P}_Z}{\mathbb{E}}[f(G(\mathbf{z}))]$$

$$-\lambda_f \cdot \underset{\substack{\mathbf{x}\sim\mathbb{P}_{\text{data}} \\ \epsilon\sim\mathbb{N}(0,I) \\ t\sim\text{Unif}(0,1)}}{\mathbb{E}} \left\| \nabla f\left(t\cdot\mathbf{x} + (1-t)\cdot G(Q^*(\mathbf{x},\epsilon))\right) - \frac{\mathbf{x} - G(Q^*(\mathbf{x},\epsilon))}{\|\mathbf{x} - G(Q^*(\mathbf{x},\epsilon))\|}\right\|^2 \quad (13)$$

$$G^* = \underset{G}{\arg\min} \underset{\substack{\mathbf{x}\sim\mathbb{P}_{\text{data}} \\ \epsilon\sim\mathbb{N}(0,I)}}{\mathbb{E}} \left[f^*(\mathbf{x}) - \lambda_{\text{mix}} \cdot f^*(G(Q^*(\mathbf{x},\epsilon)))\right] - (1-\lambda_{\text{mix}}) \cdot \underset{\mathbf{z}\sim\mathbb{P}_Z}{\mathbb{E}}[f^*(G(\mathbf{z}))] \quad (14)$$

In Eq. 12, we minimize the penalized optimal transport objective to find the optimal encoder $Q^*$ in order to produce samples from an optimal POT coupling $\tilde{\pi}^*$ (see Section 3.3). Based on this optimal encoder and the current decoder, pairs of samples from $\tilde{\pi}^*$ can be obtained by reconstructing real samples $\mathbf{x}\sim\mathbb{P}_{\text{data}}$, i.e. $(\mathbf{x}, \mathbf{y} = G(Q^*(\mathbf{x},\epsilon)))\sim\tilde{\pi}^*$.

In Eq. 13, we train the critic network $f$ to maximize the dual formulation from 4.2, while endowing it with properties of optimal potentials through the penalty term we proposed in 4.1. Minimization of this penalty term aligns the gradients (norm *and* direction) of $f$ with unit vectors that point in the direction of real samples $\mathbf{x}\sim\mathbb{P}_{\text{data}}$ from their reconstructions under $\tilde{\pi}^*$, $G(Q^*(\mathbf{x},\epsilon))$, throughout the line segments that connect them.

Finally, in Eq. 14, parameters of $G$ are learned by minimizing the dual Wasserstein distance based on the trained critic network. Ideally, each equation is solved to optimality and the optimal function is used in subsequent equations. In practice, we optimize each equation relatively well before updating the ones that depend on it. The details of the training procedure is described in more depth in Algorithm 1 in Appendix E.

The choice of hyperparameters $\lambda_{\text{mix}}$ and $\lambda_f$ depends on several factors such as the complexity of the data distribution $\mathbb{P}_{\text{data}}$, and expressivity of the function classes that are used during optimization. A crucial factor to consider is whether the optimal encoder $Q^*$ and the associated $\tilde{\pi}^*$ correspond to a valid coupling (whether it is within the set $\Pi(\mathbb{P}_{\text{data}}, \mathbb{P}_G)$), and further, how closely $\tilde{\pi}^*$ approximates an optimal coupling $\pi^*$ for the primal problem. If $\tilde{\pi}^*$ is a valid coupling or close to a valid coupling (i.e. $\mathbb{Q}_Z^* \approx \mathbb{P}_Z$), then $\lambda_{\text{mix}}$ can take a larger value within the range $[0,1]$. If $\tilde{\pi}^*$ is close to an optimal coupling, a larger $\lambda_f$ for a stronger gradient penalty may be suitable.

From the perspective of gradient-based optimization, Wasserstein Autoencoders [14] are an edge case of Primal-Dual Wasserstein GANs when $\lambda_{\text{mix}} = 1$ and $\lambda_f \to \infty$, i.e. when the gradients of $f$ are perfectly aligned on line segments associated with $\tilde{\pi}^*$. Therefore, hyperparameter pair $(\lambda_{\text{mix}}, \lambda_f)$ provide us with a space of generative modeling solutions that transition smoothly from emphasizing auto-encoding of the data distribution and stronger mode coverage with an approximately optimal coupling $\tilde{\pi}^*$, as in WAEs, to emphasizing generated sample quality ($\lambda_{\text{mix}} \to 0, \lambda_f < \infty$) and a closer approximation to optimal couplings (compared to $\tilde{\pi}^*$) through a learned and gradient regularized critic network.

## 6 Empirical Results

In this section, we empirically evaluate the proposed model. We would like to test if PD-WGANs can simultaneously achieve (i) accurate reconstructions of data points, (ii) reasonable geometry of the latent manifold, and (iii) random samples of good (visual) quality that avoids averaging effects seen in VAEs and WAEs. Importantly, the model should generalize well; requirements (i) and (ii) should be met on both training and test data. Further, we quantitatively and qualitatively compare our algorithm with Wasserstein Autoencoders (with GAN based regularizer for better quality samples) and WGAN-GPs on Inception and Frèchet Inception Distance (FID) scores. Due to limited space, we summarize our main results in this section and refer the reader to Appendix F for additional results.

We use the DC-GAN architecture as implemented in [15] for the decoder and the critic networks, which is commonly used in recent work on GANs and WAEs. We use a network similar to the critic network for the encoder sampler and use (batch) normalization only in the decoder.

**Gradient Vector Penalty** In Appendix F, Fig. 3 (middle-right), we show that the penalty term in Eq. 8 for WGAN-GPs and the penalty term in Eq. 9 for PD-WGANs, have distinct effects on the learned critic function, i.e. minimizing one does not minimize the other during training. This empirically shows that the two regularizers guide the critic network to two distinct regions in the function space.

| | **CIFAR-10** | | **CUB Birds** | | **Oxford Flowers** | |
|---|---|---|---|---|---|---|
| **Model** | FID | IS | FID | IS | FID | IS |
| WAE-GAN | 87.7 | 4.18 ± 0.04 | 143.3 | 3.42 ± 0.04 | 145.9 | 2.30 ± 0.01 |
| WGAN-GP | 34.4 | 6.58 ± 0.06 | 70.4 | 4.51 ± 0.04 | 98.7 | 3.42 ± 0.04 |
| PD-WGAN ($\lambda_{mix} = 0$) | **33.0** | **6.70 ± 0.09** | **68.6** | **4.65 ± 0.04** | **84.9** | **3.75 ± 0.04** |

Table 1: FID and Inception scores (IS) on different datasets when using DC-GAN as implemented in [15].

**Sample Quality** In Table 1, we compare our algorithm (using $\lambda_{mix} = 0$, which emphasizes sample quality) against WAEs with GAN based regularization (for better sample quality compared to WAE-MMDs) and WGAN-GPs, in terms of FID and Inception scores on three different vision datasets. PD-WGAN performs better from both baselines across all datasets. Since the regularization of the critic is the only difference between PD-WGAN and WGAN-GP in this case ($\lambda_{mix} = 0$), our results favor the gradient *vector* regularization we proposed compared to the gradient *norm* regularization used in WGAN-GPs for the critic network, as is theoretically motivated in this work. In Appendix F, Fig. 4, we provide random samples from WGAN-GPs and PD-WGANs for a visual comparison.

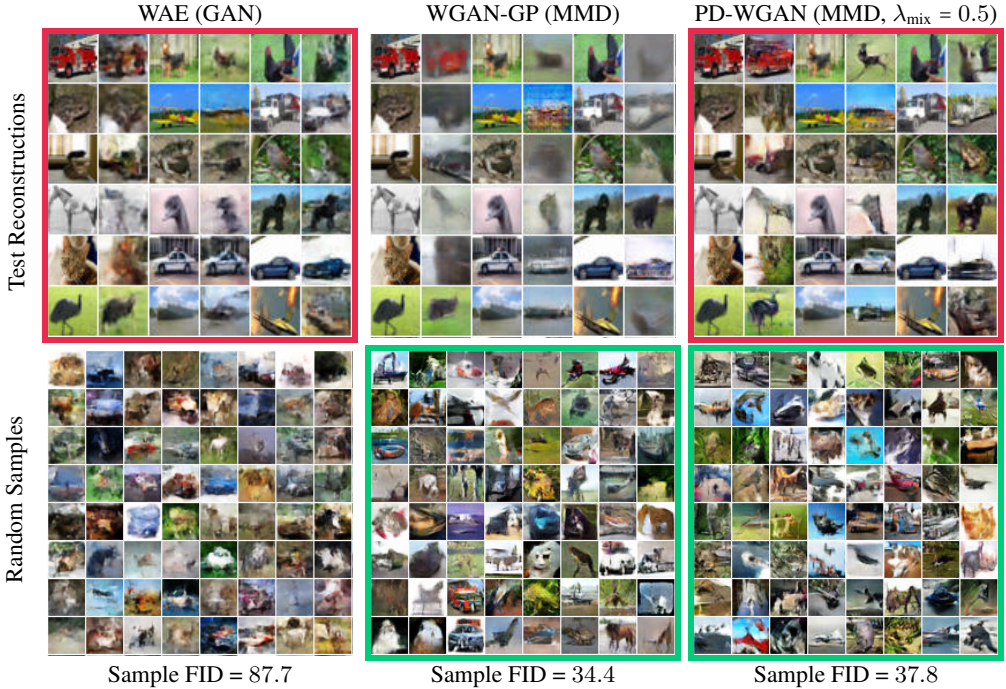

Table 2: Comparison of test reconstructions and sample quality on CIFAR-10 dataset.

**Reconstructions and Latent Manifold** In Table 2, we visually compare PD-WGANs ($\lambda_{mix} = 0.5$) to WAEs and WGAN-GPs in terms of quality of test reconstructions and quality of samples. In Appendix F, Table 3, we also show the latent manifolds learned by each algorithm by interpolating real samples across latent space. The encoder for WGAN-GP is trained with the same objective as the encoder of PD-WGANs, and does not have any effect on the training of its generative model.

Our results show that PD-WGANs have a much better generalization performance in terms of mode coverage and reconstruction quality for test samples compared to WGAN-GPs while suffering a minor degradation of sample quality in terms of FID scores. The quality gap between $\lambda_{mix} = 0$ (FID = 33.0) and $\lambda_{mix} = 0.5$ (FID = 37.8) is related to the limited capacity of the generative model and may be reduced with models with higher fidelity, as we will explore in future work.

**Future Work** We will investigate the generalization performance of these algorithms in more depth in the future, over a larger selection of datasets, architectures and more expressive generative models.

# Appendix A    Proofs for Section 4.1: Couplings and Potentials at Optimality

**Corollary 1.** *Let $(\mathcal{X}, d(\mathbf{x}, \mathbf{y}) = \|\mathbf{x} - \mathbf{y}\|)$ be a compact metric space and the unit transport cost is the distance in this metric space, $c(\mathbf{x}, \mathbf{y}) = d(\mathbf{x}, \mathbf{y}) = \|\mathbf{x} - \mathbf{y}\|$. Given $\pi^*$ is an optimal coupling for the primal and $f^*$ is an optimal potential for the dual, if $(\mathbf{x}, \mathbf{y}) \in \mathfrak{supp}\,\pi^*$ with $\mathbf{x}$ and $\mathbf{y}$ distinct, $\mathbf{x}_t = (1-t) \cdot \mathbf{x} + t \cdot \mathbf{y}$ with $0 \le t \le 1$, and $f^*$ is differentiable at $\mathbf{x}_t$, then, $\nabla f^*(\mathbf{x}_t) = {}^{\mathbf{x} - \mathbf{y}}/\|\mathbf{x} - \mathbf{y}\|$ is satisfied $\pi^*$-almost surely, i.e.*
$$\mathbb{E}_{\substack{(\mathbf{x}, \mathbf{y}) \sim \pi^* \\ t \sim \mathrm{Unif}[0,1]}} \left[ \mathbb{1}\left[ \nabla f^*(\mathbf{x}_t) = {}^{\mathbf{x} - \mathbf{y}}/\|\mathbf{x} - \mathbf{y}\| \right] \right] = 1.$$

*Proof:*

*Throughout, we assume $(\mathbf{x}, \mathbf{y})$ is an element of $\mathfrak{supp}\,\pi^*$ with $\mathbf{x}$ and $\mathbf{y}$ distinct, i.e. $\mathbf{x} \ne \mathbf{y}$.*

(i) $\frac{\mathbf{x}_t - \mathbf{y}}{\|\mathbf{x}_t - \mathbf{y}\|} = \frac{\mathbf{x} - \mathbf{y}}{\|\mathbf{x} - \mathbf{y}\|}$ *for all $t \in [0, 1)$.*

*Using the definition $\mathbf{x}_t = (1 - t) \cdot \mathbf{x} + t \cdot \mathbf{y}$,*
$$\frac{\mathbf{x}_t - \mathbf{y}}{\|\mathbf{x}_t - \mathbf{y}\|} = \frac{(1-t) \cdot (\mathbf{x} - \mathbf{y})}{\|(1-t) \cdot (\mathbf{x} - \mathbf{y})\|} = \frac{\mathbf{x} - \mathbf{y}}{\|\mathbf{x} - \mathbf{y}\|}$$

(ii) $-|t - t'| \cdot \|\mathbf{x} - \mathbf{y}\| \le f^*(\mathbf{x}_t) - f^*(\mathbf{x}'_t) \le |t - t'| \cdot \|\mathbf{x} - \mathbf{y}\|$ *for all $t, t' \in [0, 1]$.*

*Using the fact that all potentials are Lipschitz bounded, $\|f^*\|_L \le 1$,*
$$-\|\mathbf{x}_t - \mathbf{x}_{t'}\| \le f^*(\mathbf{x}_t) - f^*(\mathbf{x}'_t) \le \|\mathbf{x}_t - \mathbf{x}_{t'}\|$$
$$-\|(t - t') \cdot (\mathbf{x} - \mathbf{y})\| \le f^*(\mathbf{x}_t) - f^*(\mathbf{x}'_t) \le \|(t - t') \cdot (\mathbf{x} - \mathbf{y})\|$$
$$-|t - t'| \cdot \|\mathbf{x} - \mathbf{y}\| \le f^*(\mathbf{x}_t) - f^*(\mathbf{x}'_t) \le |t - t'| \cdot \|\mathbf{x} - \mathbf{y}\|$$

(iii) $f^*(\mathbf{x}_t) = f^*(\mathbf{y}) + (1 - t) \cdot \|\mathbf{x} - \mathbf{y}\| = f^*(\mathbf{y}) + \|\mathbf{x}_t - \mathbf{y}\|$ *for all $t \in [0, 1]$.*

*Using Theorem 1 (ii) and (ii) above to upper bound,*
$$\|\mathbf{x} - \mathbf{y}\| = f^*(\mathbf{x}) - f^*(\mathbf{y}) = f^*(\mathbf{x}_0) - f^*(\mathbf{x}_1)$$
$$= f^*(\mathbf{x}_0) - f^*(\mathbf{x}_t) + f^*(\mathbf{x}_t) - f^*(\mathbf{x}_1)$$
$$\le |t| \cdot \|\mathbf{x} - \mathbf{y}\| + |1 - t| \cdot \|\mathbf{x} - \mathbf{y}\| = \|\mathbf{x} - \mathbf{y}\|$$

*Since the l.h.s and the r.h.s of the inequality are identical, inequalities are in fact equalities.*
$$f^*(\mathbf{x}_0) - f^*(\mathbf{x}_t) = f^*(\mathbf{x}) - f^*(\mathbf{x}_t) = |t| \cdot \|\mathbf{x} - \mathbf{y}\|$$
$$f^*(\mathbf{x}_t) - f^*(\mathbf{x}_1) = f^*(\mathbf{x}_t) - f^*(\mathbf{y}) = |1 - t| \cdot \|\mathbf{x} - \mathbf{y}\|$$

*Therefore $f^*(\mathbf{x}_t) = f^*(\mathbf{y}) + (1 - t) \cdot \|\mathbf{x} - \mathbf{y}\|$. Since $(1 - t) \cdot \|\mathbf{x} - \mathbf{y}\| = \|\mathbf{x}_t - \mathbf{y}\|$, $f^*(\mathbf{x}_t) = f^*(\mathbf{y}) + \|\mathbf{x}_t - \mathbf{y}\|$, where $f^*(\mathbf{y})$ is a constant for a given line segment.*

(iv) *Directional derivative along $\frac{\mathbf{x}_t - \mathbf{y}}{\|\mathbf{x}_t - \mathbf{y}\|}$ is 1, i.e. $\lim_{h \to 0} \frac{f^*\left(\mathbf{x}_t + h \cdot \frac{\mathbf{x}_t - \mathbf{y}}{\|\mathbf{x}_t - \mathbf{y}\|}\right) - f^*(\mathbf{x}_t)}{h} = 1$.*

$$f^*\left(\mathbf{x}_t + h \cdot \frac{\mathbf{x}_t - \mathbf{y}}{\|\mathbf{x}_t - \mathbf{y}\|}\right) - f^*(\mathbf{x}_t) = \left\|\mathbf{x}_t + h \cdot \frac{\mathbf{x}_t - \mathbf{y}}{\|\mathbf{x}_t - \mathbf{y}\|} - \mathbf{y}\right\| - \|\mathbf{x}_t - \mathbf{y}\|$$
$$= \left(1 + \frac{h}{\|\mathbf{x}_t - \mathbf{y}\|}\right)\|\mathbf{x}_t - \mathbf{y}\| - \|\mathbf{x}_t - \mathbf{y}\| = h$$

*Therefore $\lim_{h \to 0} \frac{f^*\left(\mathbf{x}_t + h \cdot \frac{\mathbf{x}_t - \mathbf{y}}{\|\mathbf{x}_t - \mathbf{y}\|}\right) - f^*(\mathbf{x}_t)}{h} = \lim_{h \to 0} \frac{h}{h} = 1$*

*Using (iv), the gradient $\nabla f^*(\mathbf{x}_t)$ can be decomposed as $\nabla f^*(\mathbf{x}_t) = 1 \cdot {}^{\mathbf{x}_t - \mathbf{y}}/\|\mathbf{x}_t - \mathbf{y}\| + \mathbf{v}_\perp$ where ${}^{\mathbf{x}_t - \mathbf{y}}/\|\mathbf{x}_t - \mathbf{y}\| \cdot \mathbf{v}_\perp = 0$, i.e. $\mathbf{v}_\perp$ represents the component of $\nabla f^*(\mathbf{x}_t)$ that is perpendicular to ${}^{\mathbf{x}_t - \mathbf{y}}/\|\mathbf{x}_t - \mathbf{y}\|$. Then, $\|\nabla f^*(\mathbf{x}_t)\| = \sqrt{1 + \|\mathbf{v}_\perp\|^2}$. Since the norm of the gradient is bounded as $\|\nabla f^*(x)\| \le 1$ (all potentials $f$ are Lipschitz bounded), the perpendicular component must be $\mathbf{v}_\perp = \vec{0}$. Therefore, in combination with (i), $\nabla f^*(\mathbf{x}_t) = \frac{\mathbf{x}_t - \mathbf{y}}{\|\mathbf{x}_t - \mathbf{y}\|} = \frac{\mathbf{x} - \mathbf{y}}{\|\mathbf{x} - \mathbf{y}\|}$.*

*This completes the proof.* □

**Monge-Mather shortening principle, [7].** *No optimal coupling $\pi^*$ can have on its support two distinct pairs $(\mathbf{x}_1, \mathbf{y}_1)$ and $(\mathbf{x}_2, \mathbf{y}_2)$ such that the line segments associated with them intersect at an interior point, $\mathbf{m} = (1-t_1) \cdot \mathbf{x}_1 + t_1 \cdot \mathbf{y}_1 = (1-t_2) \cdot \mathbf{x}_2 + t_2 \cdot \mathbf{y}_2$ where $t_1, t_2 \in (0, 1)$.*

*Assume $\pi$ is a coupling and $\mathbf{m}$ exists for two distinct pairs on its support as given. Then, coupling $\pi^+$ that swaps the pairs $(\mathbf{x}_1, \mathbf{y}_1)$ and $(\mathbf{x}_2, \mathbf{y}_2)$ with $(\mathbf{x}_1, \mathbf{y}_2)$ and $(\mathbf{x}_2, \mathbf{y}_1)$, has a strictly lower aggregate transport cost than $\pi$, since (using the triangle inequality, see Figure 2),*

$$\|\mathbf{x}_1 - \mathbf{y}_2\| + \|\mathbf{x}_2 - \mathbf{y}_1\| < \|\mathbf{x}_1 - \mathbf{y}_1\| + \|\mathbf{x}_2 - \mathbf{y}_2\|.$$

*This breaks the c-cyclic monotonicity property of optimal couplings [7]. Therefore, $\pi$ cannot be an optimal coupling.*

**Proposition 1.** *Given a neighborhood exists around any point $\mathbf{y} \in \mathrm{supp}\,\mathbb{P}_G$ along $\pm(\mathbf{x}_2-\mathbf{x}_1)/\|\mathbf{x}_2-\mathbf{x}_1\|$ for any two points $\mathbf{x}_1, \mathbf{x}_2 \in \mathrm{supp}\,\mathbb{P}_{\mathrm{data}}$, then $\mathrm{supp}\,\pi^\otimes \neq \mathrm{supp}\,\pi^*$.*

*Proof:*

*Given $\mathbf{x}_1, \mathbf{x}_2 \in \mathrm{supp}\,\mathbb{P}_{\mathrm{data}}$ and there exists an $\epsilon \in [0, \infty)$ such that $\mathbf{y}_1 = \mathbf{y} + \epsilon \cdot \frac{\mathbf{x}_2-\mathbf{x}_1}{\|\mathbf{x}_2-\mathbf{x}_1\|} \in \mathrm{supp}\,\mathbb{P}_G$ and $\mathbf{y}_2 = \mathbf{y} - \epsilon \cdot \frac{\mathbf{x}_2-\mathbf{x}_1}{\|\mathbf{x}_2-\mathbf{x}_1\|} \in \mathrm{supp}\,\mathbb{P}_G$, then, the two pairs $(\mathbf{x}_1, \mathbf{y}_1), (\mathbf{x}_2, \mathbf{y}_2) \in \mathrm{supp}\,\pi^\otimes$. Consider the line segment between $\mathbf{x}_1$ and $\mathbf{y}_1$; and the line segment between $\mathbf{x}_2$ and $\mathbf{y}_2$. Given $t_1 = t_2 = \frac{\|\mathbf{x}_2-\mathbf{x}_1\|}{2\epsilon + \|\mathbf{x}_2-\mathbf{x}_1\|} \in (0, 1)$, these line segments meet at an interior point $\mathbf{m}$:*

$$\mathbf{m} = (1-t_1) \cdot \mathbf{x}_1 + t_1 \cdot \mathbf{y}_1 = (1-t_2) \cdot \mathbf{x}_2 + t_2 \cdot \mathbf{y}_2$$

$$\mathbf{m} = \left(1 - \frac{\|\mathbf{x}_2 - \mathbf{x}_1\|}{2\epsilon + \|\mathbf{x}_2 - \mathbf{x}_1\|}\right) \cdot \mathbf{x}_1 + \frac{\|\mathbf{x}_2 - \mathbf{x}_1\|}{2\epsilon + \|\mathbf{x}_2 - \mathbf{x}_1\|} \cdot \left(\mathbf{y} + \epsilon \cdot \frac{\mathbf{x}_2 - \mathbf{x}_1}{\|\mathbf{x}_2 - \mathbf{x}_1\|}\right)$$

$$\mathbf{m} = \frac{2\epsilon \cdot \mathbf{x}_1 + \|\mathbf{x}_2 - \mathbf{x}_1\| \cdot \mathbf{y} + \epsilon \cdot (\mathbf{x}_2 - \mathbf{x}_1)}{2\epsilon + \|\mathbf{x}_2 - \mathbf{x}_1\|}$$

*Since no optimal coupling can have on its support such a point (according to Monge-Mather shortening principle), $\mathrm{supp}\,\pi^\otimes \neq \mathrm{supp}\,\pi^*$.*

*This completes the proof.* □

**Proposition 2.** *Given $(\mathbf{x}, \mathbf{y}) \in \mathrm{supp}\,\pi^*$, and $f^*$ is differentiable at $\mathbf{x}$, then $\mathrm{supp}\,\pi^*_{Y|X=\mathbf{x}}$ is confined to (a subset of) the half line given by $\mathbf{x} + t \cdot (\mathbf{y}-\mathbf{x}/\|\mathbf{y}-\mathbf{x}\|)$ where $t \in [0, \infty)$.*

*Proof:*

*Using Corollary 1, gradient of $f^*$ at $\mathbf{x}$ is given as $\nabla f^*(\mathbf{x}_t) = \frac{\mathbf{x}-\mathbf{y}}{\|\mathbf{x}-\mathbf{y}\|}$. If there exists $\mathbf{y}' \in \mathrm{supp}\,\pi^*_{Y|X=\mathbf{x}}$, then $(\mathbf{x}, \mathbf{y}') \in \mathrm{supp}\,\pi^*$ as well. Then, for any such $\mathbf{y}'$, gradient of $f^*$ at $\mathbf{x}$ must also be $\nabla f^*(\mathbf{x}_t) = \frac{\mathbf{x}-\mathbf{y}'}{\|\mathbf{x}-\mathbf{y}'\|}$. Therefore, $\frac{\mathbf{x}-\mathbf{y}'}{\|\mathbf{x}-\mathbf{y}'\|} = \frac{\mathbf{x}-\mathbf{y}}{\|\mathbf{x}-\mathbf{y}\|}$. This can only be true if any such $\mathbf{y}'$ is an element of the half line given by $\mathbf{x} + t \cdot (\mathbf{y}-\mathbf{x}/\|\mathbf{y}-\mathbf{x}\|)$ where $t \in [0, \infty)$. Then, $\mathrm{supp}\,\pi^*_{Y|X=\mathbf{x}}$ cannot lie outside this half line.*

*This completes the proof.* □

**Corollary 2.** *Given there exists a point $\mathbf{x} \in \mathrm{supp}\,\mathbb{P}_{\mathrm{data}}$ where $f^*$ is differentiable at $\mathbf{x}$ and $\mathrm{supp}\,\mathbb{P}_G$ does not lie on a single line, then $\mathrm{supp}\,\pi^\otimes \neq \mathrm{supp}\,\pi^*$.*

*Proof:* *Given $\mathbf{x} \in \mathrm{supp}\,\mathbb{P}_{\mathrm{data}}$, $\mathbf{y} \in \mathrm{supp}\,\mathbb{P}_G$, then $(\mathbf{x}, \mathbf{y}) \in \mathrm{supp}\,\pi^\otimes$. If $\mathrm{supp}\,\mathbb{P}_G$ does not lie on a single line, then $\mathrm{supp}\,\pi^\otimes_{Y|X=\mathbf{x}}$ does not lie on a single line since $\pi^\otimes_{Y|X=\mathbf{x}} = \mathbb{P}_G$. Since for any $\pi^*$, $\mathrm{supp}\,\pi^*_{Y|X=\mathbf{x}}$ is a subset of the half line given by $\mathbf{x} + t \cdot (\mathbf{y}-\mathbf{x}/\|\mathbf{y}-\mathbf{x}\|)$ where $t \in [0, \infty)$, $\mathrm{supp}\,\pi^\otimes_{Y|X=\mathbf{x}} \neq \mathrm{supp}\,\pi^*_{Y|X=\mathbf{x}}$. Therefore, $\mathrm{supp}\,\pi^\otimes \neq \mathrm{supp}\,\pi^*$.*

*This completes the proof.* □

## Appendix B  Coupling Based Reformulation of the Dual Problem

Assume $\pi$ is a coupling between $\mathbb{P}_{\text{data}}$ and $\mathbb{P}_G$, i.e. $\pi \in \Pi(\mathbb{P}_{\text{data}}, \mathbb{P}_G)$. Then, the dual optimal transport problem can be rewritten as follows:

$$\begin{aligned}
\text{OT}_d(\mathbb{P}_{\text{data}}, \mathbb{P}_G) &= \sup_{f \in \mathcal{F}_{\text{Lip}}} \mathbb{E}_{\mathbf{x} \sim \mathbb{P}_{\text{data}}}[f(\mathbf{x})] - \mathbb{E}_{\mathbf{y} \sim \mathbb{P}_G}[f(\mathbf{y})] \\
&= \sup_{f \in \mathcal{F}_{\text{Lip}}} \mathbb{E}_{\mathbf{x} \sim \mathbb{P}_{\text{data}}}[f(\mathbf{x})] - \left[\lambda_{\text{mix}} \cdot \mathbb{E}_{(\mathbf{x},\mathbf{y}) \sim \pi}[f(\mathbf{y})] + (1 - \lambda_{\text{mix}}) \cdot \mathbb{E}_{\mathbf{y} \sim \mathbb{P}_G}[f(\mathbf{y})]\right] \\
&= \sup_{f \in \mathcal{F}_{\text{Lip}}} \mathbb{E}_{\mathbf{x} \sim \mathbb{P}_{\text{data}}}[f(\mathbf{x})] + \mathbb{E}_{\mathbf{x} \sim \mathbb{P}_{\text{data}}}\left[-\lambda_{\text{mix}} \cdot \mathbb{E}_{\mathbf{y} \sim \pi_{Y|X=\mathbf{x}}}[f(\mathbf{y})]\right] - (1 - \lambda_{\text{mix}}) \cdot \mathbb{E}_{\mathbf{y} \sim \mathbb{P}_G}[f(\mathbf{y})] \\
&= \sup_{f \in \mathcal{F}_{\text{Lip}}} \mathbb{E}_{\mathbf{x} \sim \mathbb{P}_{\text{data}}}\left[f(\mathbf{x}) - \lambda_{\text{mix}} \cdot \mathbb{E}_{\mathbf{y} \sim \pi_{Y|X=\mathbf{x}}}[f(\mathbf{y})]\right] - (1 - \lambda_{\text{mix}}) \cdot \mathbb{E}_{\mathbf{y} \sim \mathbb{P}_G}[f(\mathbf{y})] \\
&= \sup_{f \in \mathcal{F}_{\text{Lip}}} \mathbb{E}_{(\mathbf{x},\mathbf{y}) \sim \pi}[f(\mathbf{x}) - \lambda_{\text{mix}} \cdot f(\mathbf{y})] - (1 - \lambda_{\text{mix}}) \cdot \mathbb{E}_{\mathbf{z} \sim \mathbb{P}_Z}[f(G(\mathbf{z}))]
\end{aligned}$$

Similar to the reparameterization introduced in Section 3.1, when the generative model $\mathbb{P}_G$ is defined through a prior over latent codes and a deterministic decoder, we can replace couplings with conditional distributions $\mathbb{Q}_{Z|X}$ with aggregated posteriors equal to the prior $\mathbb{P}_Z$. Given $\mathbb{Q}_Z = \mathbb{P}_Z$, the dual OT distance is

$$\text{OT}_d(\mathbb{P}_{\text{data}}, \mathbb{P}_G) = \sup_{f \in \mathcal{F}_{\text{Lip}}} \mathbb{E}_{\mathbf{x} \sim \mathbb{P}_{\text{data}}}\left[f(\mathbf{x}) - \lambda_{\text{mix}} \cdot \mathbb{E}_{\mathbf{z} \sim \mathbb{Q}_{Z|X=\mathbf{x}}}[f(G(\mathbf{z}))]\right] - (1 - \lambda_{\text{mix}}) \cdot \mathbb{E}_{\mathbf{z} \sim \mathbb{P}_Z}[f(G(\mathbf{z}))].$$

## Appendix C  Proofs for Section 4.2: Auto-encoding Critic Networks

**Proposition 3.** *Assume $\theta$ is a set of variables that parameterizes the decoder $G$, $\mathbf{x} \sim \mathbb{P}_{\text{data}}$ is any given data sample and $f \in \mathcal{F}_{\text{Lip}}$ is a (possibly sub-optimal) potential function that $\mathbb{Q}_{Z|X=\mathbf{x}}$-almost surely satisfies $\nabla f(G(\mathbf{z})) = \mathbf{x} - G(\mathbf{z})/\|\mathbf{x} - G(\mathbf{z})\|$ where $\mathbb{Q}_{Z|X}$ is any conditional distribution. Then, the following statement is true:*

$$\nabla_\theta \mathbb{E}_{\mathbf{x} \sim \mathbb{P}_{\text{data}}}\left[f(\mathbf{x}) - \lambda_{\text{mix}} \cdot \mathbb{E}_{\mathbf{z} \sim \mathbb{Q}_{Z|X=\mathbf{x}}}[f(G(\mathbf{z}))]\right] = \lambda_{\text{mix}} \cdot \nabla_\theta \mathbb{E}_{\mathbf{x} \sim \mathbb{P}_{\text{data}}}\left[\mathbb{E}_{\mathbf{z} \sim \mathbb{Q}_{Z|X=\mathbf{x}}}\|\mathbf{x} - G(\mathbf{z})\|\right].$$

*Proof:*

*We are given* $\mathbb{E}_{\substack{\mathbf{x} \sim \mathbb{P}_{\text{data}} \\ \mathbf{z} \sim \mathbb{Q}_{Z|X=\mathbf{x}}}}\left[\mathbb{1}\left[\nabla f(G(\mathbf{z})) = \frac{\mathbf{x} - G(\mathbf{z})}{\|\mathbf{x} - G(\mathbf{z})\|}\right]\right] = 1.$ *Therefore the following is true also,*

$\mathbb{E}_{\substack{\mathbf{x} \sim \mathbb{P}_{\text{data}} \\ \mathbf{z} \sim \mathbb{Q}_{Z|X=\mathbf{x}}}}\left[\mathbb{1}\left[-\nabla_\theta G(\mathbf{z}) \cdot \nabla f(G(\mathbf{z})) = -\nabla_\theta G(\mathbf{z}) \cdot \frac{\mathbf{x} - G(\mathbf{z})}{\|\mathbf{x} - G(\mathbf{z})\|}\right]\right] = 1.$ *Integrals of almost-surely equal functions are equal as well under the same measure. We use this fact in step 4.*

$$\begin{aligned}
\nabla_\theta \mathbb{E}_{\mathbf{x} \sim \mathbb{P}_{\text{data}}}\left[f(\mathbf{x}) - \lambda_{\text{mix}} \cdot \mathbb{E}_{\mathbf{z} \sim \mathbb{Q}_{Z|X=\mathbf{x}}}[f(G(\mathbf{z}))]\right] &= \lambda_{\text{mix}} \cdot \mathbb{E}_{\mathbf{x} \sim \mathbb{P}_{\text{data}}}\left[\mathbb{E}_{\mathbf{z} \sim \mathbb{Q}_{Z|X=\mathbf{x}}}[\nabla_\theta[-f(G(\mathbf{z}))]]\right] \\
&= \lambda_{\text{mix}} \cdot \mathbb{E}_{\mathbf{x} \sim \mathbb{P}_{\text{data}}}\left[\mathbb{E}_{\mathbf{z} \sim \mathbb{Q}_{Z|X=\mathbf{x}}}[-\nabla_\theta G(\mathbf{z}) \cdot \nabla_{G(\mathbf{z})}[f(G(\mathbf{z}))]]\right] \\
&= \lambda_{\text{mix}} \cdot \mathbb{E}_{\mathbf{x} \sim \mathbb{P}_{\text{data}}}\left[\mathbb{E}_{\mathbf{z} \sim \mathbb{Q}_{Z|X=\mathbf{x}}}[-\nabla_\theta G(\mathbf{z}) \cdot \nabla f(G(\mathbf{z}))]\right] \\
\text{[Step 4]} \quad &= \lambda_{\text{mix}} \cdot \mathbb{E}_{\mathbf{x} \sim \mathbb{P}_{\text{data}}}\left[\mathbb{E}_{\mathbf{z} \sim \mathbb{Q}_{Z|X=\mathbf{x}}}\left[-\nabla_\theta G(\mathbf{z}) \cdot \frac{\mathbf{x} - G(\mathbf{z})}{\|\mathbf{x} - G(\mathbf{z})\|}\right]\right] \\
&= \lambda_{\text{mix}} \cdot \mathbb{E}_{\mathbf{x} \sim \mathbb{P}_{\text{data}}}\left[\mathbb{E}_{\mathbf{z} \sim \mathbb{Q}_{Z|X=\mathbf{x}}}\left[\nabla_\theta G(\mathbf{z}) \cdot \nabla_{G(\mathbf{z})}\|\mathbf{x} - G(\mathbf{z})\|\right]\right] \\
&= \lambda_{\text{mix}} \cdot \mathbb{E}_{\mathbf{x} \sim \mathbb{P}_{\text{data}}}\left[\mathbb{E}_{\mathbf{z} \sim \mathbb{Q}_{Z|X=\mathbf{x}}}\left[\nabla_\theta \|\mathbf{x} - G(\mathbf{z})\|\right]\right]
\end{aligned}$$

$$\nabla_\theta \mathbb{E}_{\mathbf{x} \sim \mathbb{P}_{\text{data}}}\left[f(\mathbf{x}) - \lambda_{\text{mix}} \cdot \mathbb{E}_{\mathbf{z} \sim \mathbb{Q}_{Z|X=\mathbf{x}}}[f(G(\mathbf{z}))]\right] = \lambda_{\text{mix}} \cdot \nabla_\theta \mathbb{E}_{\mathbf{x} \sim \mathbb{P}_{\text{data}}}\left[\mathbb{E}_{\mathbf{z} \sim \mathbb{Q}_{Z|X=\mathbf{x}}}\|\mathbf{x} - G(\mathbf{z})\|\right]$$

*This completes the proof.* □

# Appendix D  Divergences and Metrics in Probability Space

A commonly used family of statistical divergences are called $f$-divergences, $\mathfrak{D}_f(\mu\|\nu) = \mathbb{E}_\mu[f(d\nu/d\mu)]$, such as the popular Kullback-Leibler (KL) divergence used in maximum likelihood estimation. Although popular, $f$-divergences compare relative probabilities between measures for *identical* outcomes (i.e. the class of divergences invariant to invertible transformations [16]) and do not take into account how close two *different* outcomes may be in the geometry of the sample space, which is desirable in certain cases [19, 20].

A related limitation of this relative probability based approach is that the divergence may saturate or go to infinity if the two measures are not absolutely continuous with respect to each other, e.g. when a set of outcomes has a non-zero measure under one distribution but not the other. A mismatch between supports is exceedingly likely in high dimensional spaces when dealing with distributions that have intrinsically lower-dimensional supports. This is especially problematic for gradient based optimization methods since a saturated or maxed-out discrepancy does not provide useful gradients for training.

These limitations motivate the use of other flexible families of divergences that induce weaker topologies [5] such as *Integral Probability Metrics* (IPMs), $\mathfrak{D}_\mathcal{F}(\mu\|\nu) = \sup_{f\in\mathcal{F}} |\mathbb{E}_\mu[f] - \mathbb{E}_\nu[f]|$, which can take into account the underlying geometry in the sample space and can be applied to measures with non-overlapping supports. Based on the choice of the function class[5] $\mathcal{F}$, examples of this family of metrics include Total Variation (TV) distance, Dudley metric, Cramer distance, Maximum Mean Discrepancy (MMD), and *Wasserstein* distances.

# Appendix E  Algorithm

**Algorithm 1** Primal-Dual Wasserstein GAN for $\lambda_{\text{mix}} = 0$. All experiments in the paper use the default values: $\lambda_f = 1, \lambda_Z = 10, n_{\text{encoder}} = 1, n_{\text{critic}} = 5, \alpha = 10^{-4}, \beta_1 = 0.5, \beta_2 = 0.9, m = 50$.

**Require:** The encoder penalty coefficient $\gamma$, the critic penalty coefficient $\lambda$, the number of encoder iterations per critic iteration $n_{\text{encoder}}$, the number of critic iterations per generator iteration $n_{\text{critic}}$, Adam hyperparameters $\alpha, \beta_1, \beta_2$, the mini-batch size $m$.

**Require:** Initial encoder parameters $\phi_0$, initial critic parameters $\omega_0$, initial generator parameters $\theta_0$.

1: **while** $\theta_0$ is not converged **do**
2:     Sample $\{\mathbf{x}^{(l)}\}_{l=1}^m \sim \mathbb{P}_{\text{data}}$, a batch from the empirical data distribution.
3:     Sample $\{\mathbf{z}^{(l)}\}_{l=1}^m \sim \mathbb{P}_Z$, a batch from the prior distribution.
4:     Sample $\{\epsilon^{(l)}\}_{l=1}^m \sim \mathbb{N}(0, I)$, a batch of random noise components.
5:     Sample $\{t^{(l)}\}_{l=1}^m \sim \text{Unif}[0,1]$, a batch of uniform samples.
6:
7:     **for** $i = 1, \ldots, n_{\text{critic}}$ **do**
8:         **for** $j = 1, \ldots, n_{\text{encoder}}$ **do**
9:             $K_{\mathbb{P}_Z} = \frac{1}{m(m-1)} \sum_{l=1}^m \sum_{l' \neq l} k(\mathbf{z}^{(l)}, \mathbf{z}^{(l')})$
10:            $K_Q = \frac{1}{m(m-1)} \sum_{l=1}^m \sum_{l' \neq l} k(Q(\mathbf{x}^{(l)}, \epsilon^{(l)}), Q(\mathbf{x}^{(l')}, \epsilon^{(l')}))$
11:            $K_{\mathbb{P}_Z,Q} = \frac{1}{m^2} \sum_{l=1}^m \sum_{l'=1}^m k(\mathbf{z}^{(l)}, Q(\mathbf{x}^{(l')}, \epsilon^{(l')}))$
12:
13:            $J_Q^{\text{Penalty}} = -(K_{\mathbb{P}_Z} + K_Q - 2K_{\mathbb{P}_Z,Q})$     ▷ MMD based penalty.
14:            $J_Q = -\frac{1}{m} \sum_{l=1}^m \|\mathbf{x}^{(l)} - G(Q(\mathbf{x}^{(l)}, \epsilon^{(l)}))\|$
15:            $\phi \leftarrow \text{Adam}(\nabla_\phi(J_Q + \gamma \cdot J_Q^{\text{Penalty}}), \phi, \alpha, \beta_1, \beta_2)$
16:         **end for**
17:         $J_f^{\text{Penalty}} = -\frac{1}{m} \sum_{l=1}^m \left\| \nabla\psi\big(t^{(l)} \cdot \mathbf{x}^{(l)} + (1-t^{(l)}) \cdot G(Q(\mathbf{x}^{(l)}, \epsilon^{(l)}))\big) - \frac{\mathbf{x}^{(l)} - G(Q(\mathbf{x}^{(l)}, \epsilon^{(l)}))}{\|\mathbf{x}^{(l)} - G(Q(\mathbf{x}^{(l)}, \epsilon^{(l)}))\|} \right\|^2$
18:         $J_f = \frac{1}{m} \sum_{l=1}^m f(\mathbf{x}^{(l)}) - f(G(\mathbf{z}^{(l)}))$
19:         $\omega \leftarrow \text{Adam}(\nabla_\omega(J_f + \lambda \cdot J_f^{\text{Penalty}}), \omega, \alpha, \beta_1, \beta_2)$
20:     **end for**
21:     $J_G = -\frac{1}{m} \sum_{l=1}^m \|\mathbf{x}^{(l)} - G(Q(\mathbf{x}^{(l)}, \epsilon^{(l)}))\| + \frac{1}{m} \sum_{l=1}^m \psi(G(\mathbf{z}^{(l)}))$
22:     $\theta \leftarrow \text{Adam}(\nabla_\theta(J_G), \theta, \alpha, \beta_1, \beta_2)$
23: **end while**

---

[5]A subset of real-valued, bounded, measurable functions.

## Appendix F  Additional Results

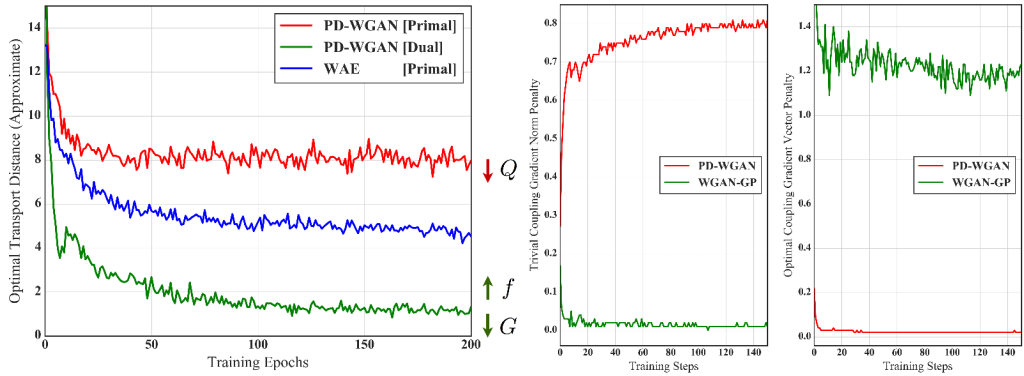

**Figure 3: Left:** Estimated primal and dual optimal transport distance during training for WAEs (primal–in blue) and PD-WGANs (primal and dual–red and green). **Middle:** Evaluation of the gradient norm penalty used in WGAN-GPs when the model being trained is WGAN-GP (green) and PD-WGAN (red). **Right:** Evaluation of the gradient vector penalty we propose for PD-WGANs in Eq. 13 when the model being trained is WGAN-GP (green) and PD-WGAN (red). In practice, minimization of one penalty does not minimize the other and each of them guides the critic network $f$ to a distinct region in the function space.

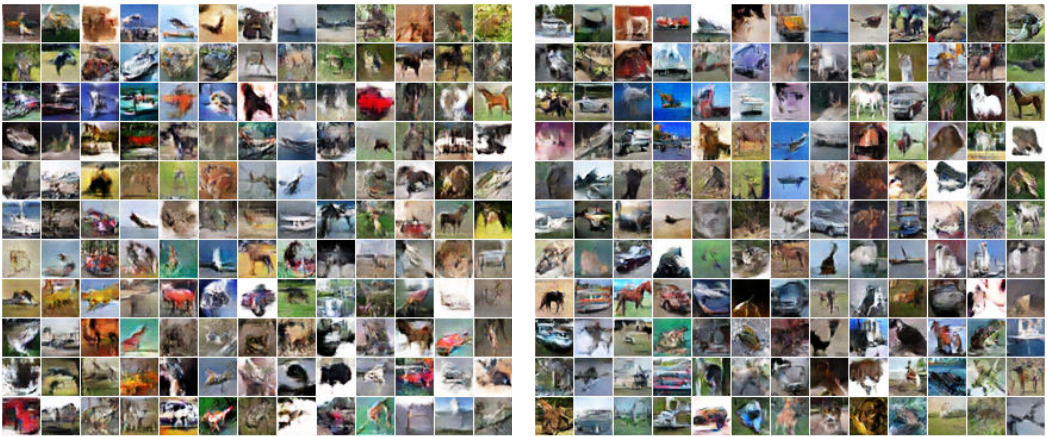

**Figure 4: Left:** Random samples from WGAN-GP when using the DC-GAN architecture as implemented in [15]. **Right:** Random samples from PD-WGAN with $\lambda_{\text{mix}} = 0$ when using the same architecture.

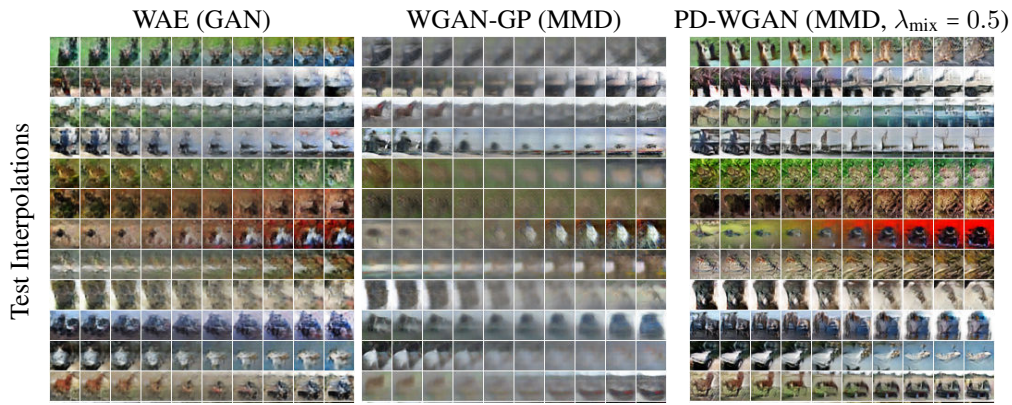

**Table 3:** Latent space interpolations for different algorithms for CIFAR-10 dataset.

14